\documentclass{isprs} 
\usepackage{subcaption} 
\usepackage{setspace}
\usepackage{geometry} 
\usepackage{epstopdf}
\usepackage[labelsep=period]{caption}  
\usepackage[british]{babel}
\usepackage{epsfig}
\usepackage{graphicx}
\usepackage{caption}
\usepackage{amsmath}
\usepackage{amssymb}
\usepackage{tabularx}
\usepackage{booktabs, makecell}
\usepackage{makecell}
\usepackage{soul}
\usepackage{hyperref}
\usepackage{array}
\newcolumntype{P}[1]{>{\centering\arraybackslash}p{#1}}
\usepackage[dvipsnames]{xcolor}
\newcommand{\lidar}{LIDAR }
\newcommand{\lidars}{LIDARs }
\newcommand{\Fone}{F$_1$ }
\newcommand{\image}{\mathrm{R}}

\DeclareMathOperator{\atantwo}{atan2}
\begin{document}

\title{Road Segmentation on low resolution \lidar point clouds for autonomous vehicles}
\author{
 Leonardo Gigli\textsuperscript{1}, B Ravi Kiran\textsuperscript{2}, Thomas Paul\textsuperscript{4}, Andres Serna\textsuperscript{3},
 Nagarjuna Vemuri\textsuperscript{4}, Beatriz Marcotegui\textsuperscript{1}, Santiago Velasco-Forero\textsuperscript{1}}

 \address{
 	\textsuperscript{1 } Center for Mathematical Morphology (CMM) - MINES ParisTech - PSL Research University\\ Fontainebleau, France - \emph{name.surname}@mines-paristech.fr\\
 	\textsuperscript{2 }NavyaTech - Paris, France - ravi.kiran@navya.tech\\
 	\textsuperscript{3 }Terra3D - Paris, France - andres.serna@terra3d.fr\\
 	\textsuperscript{4 }Independent researchers - thomaspaul582@gmail.com, nagarjuna0911@gmail.com
 }

\commission{}{} 
\workinggroup{} 
\icwg{}   

\abstract{
Point cloud datasets for perception tasks in the context of autonomous driving often rely on high resolution 64-layer Light Detection and Ranging (LIDAR) scanners. They are expensive to deploy on real-world autonomous driving sensor architectures which usually employ 16/32 layer LIDARs. We evaluate the effect of subsampling image based representations of dense point clouds on the accuracy of the road segmentation task. In our experiments the low resolution 16/32 layer \lidar point clouds are simulated by subsampling the original 64 layer data, for subsequent transformation in to a feature map in the Bird-Eye-View(BEV) and Spherical-View (SV) representations of the point cloud. We introduce the usage of the local normal vector with the LIDAR's spherical coordinates as an input channel to existing LoDNN architectures. We demonstrate that this local normal feature in conjunction with classical features not only improves performance for binary road segmentation on full resolution point clouds, but it also reduces the negative impact on the accuracy when subsampling dense point clouds as compared to the usage of classical features alone. We assess our method with several experiments on two datasets: KITTI Road-segmentation benchmark and the recently released Semantic KITTI dataset.
}


\maketitle


\section{Introduction}\label{sec:introduction}

\sloppy
Modern day \lidars are multi-layer $3$D laser scanners that enable a $3$D-surface reconstruction of large-scale environments. They provide precise range information while poorer semantic information as compared to color cameras. They are thus employed in obstacle avoidance and SLAM (Simultaneous localization and Mapping) applications. The number of layers and angular steps in elevation \& azimuth of the \lidar characterizes the spatial resolution. With the recent development in the automated driving (AD) industry the \lidar sensor industry has gained increased attention. \lidar scan-based point cloud datasets for AD such as KITTI usually were generated by high-resolution \lidar (64 layers, 1000 azimuth angle positions \cite{Fritsch2013ITSC}), referred to as a dense point cloud scans. In recent nuScenes dataset for multi-modal object detection a 32-Layer \lidars scanner has been used for acquisition \cite{nuscenes2019}. Another source of datasets are large-scale point clouds which achieve a high spatial resolution by aggregating multiple closely-spaced point clouds, aligned using the mapping vehicle's pose information obtained using GPS-GNSS based localization and orientation obtained using inertial moment units (IMUs) \cite{roynard2017parisIJRR}. Large-scale point clouds are employed in the creation of high-precision semantic map representation of environments and have been studied for different applications such as detection and segmentation of urban objects \cite{serna2014detection}. We shall focus on the scan-based point cloud datasets in our study.

Road segmentation is an essential component of the autonomous driving tasks. In complement with obstacle avoidance, trajectory planning and driving policy, it is a key real-time task to extract the drivable free space as well as determine the road topology. Recent usage and proliferation of DNNs (Deep neural networks) for various perception tasks in point clouds has opened up many interesting applications. A few applications relating to road segmentation include, binary road segmentation \cite{lodnn2017} where the goal is classify the point cloud set into road and non road $3$D points. Ground extraction \cite{velas2018cnn} regards the problem of obtaining the border between the obstacle and the ground. Finally, recent benchmark for semantic segmentation of point clouds was released with the Semantic-KITTI dataset by \cite{behley2019iccv}. In Rangenet++ \cite{milioto2019rangenet++} authors evaluate the performance of Unet \& Darknet architectures for the task of semantic segmentation on point clouds. This includes the road scene classes such as pedestrians, cars, sidewalks, vegetation, road, among others.

\subsection{Motivation \& Contributions}

We first observe that different \lidar senor configurations produce different distribution of points in the scanned $3$D point cloud. The configurations refer to, \lidar position \& orientation, the vertical field-of-view (FOV), angular resolution and thus number of layers, the elevation and azimuth angles that the lasers scan through. These differences directly affect the performance of deep learning models that learn representations for different tasks, such as semantic segmentation and object detection. Low-resolution 16 layer \lidars have been recently compared with 64 layer \lidars \cite{del2017low} to evaluate the degradation in detection accuracy especially w.r.t distance. From Table \ref{tab:lidar-configurations} we observe that the HDL-64 contains 4x more points than VLP-16. This increases the computational time \& memory requirements (GPU or CPU) to run the road segmentation algorithms. Thus, it is a computational challenge to process a large amount of points in real-time.

\begin{table*}
\centering
\begin{tabular}{|m{4.5cm}|m{3.70cm}|m{3.70cm}|m{3.70cm}|} 
  \hline
  {\textbf{\lidar}} & {\textbf{Velodyne HDL-64}} & {\textbf{Velodyne HDL-32}} & {\textbf{Velodyne VLP-16}} \\
  \hline
  Azimuth            & \makecell{[$0^{\circ}, 360^{\circ}$) \\ step $0.18^\circ$} & \makecell{[$0^{\circ}, 360^{\circ}$) \\ step $0.1^\circ-0.4^\circ$}& \makecell{[$0^{\circ}, 360^{\circ}$) \\ step $0.2^\circ$}\\
  \hline
  Elevation          & \makecell{[$-24.3^{\circ}, 2^{\circ}$]\\ step 1-32 : $1/3^\circ$ \\ step 33-64 : $1/2^\circ$} & \makecell{[$+10.67^\circ, -30.67^\circ$] \\ $1.33^\circ$ for 32 layers} & \makecell{[$-15^\circ, 15^\circ$]\\ $2^\circ$ for 16 layers}\\
  \hline
  Price (as reviewed on 2019)              & $\sim85$ k\$ & $\sim20$ k\$ & $\sim4$ k\$ \\
  \hline
  Effective Vertical FOV            & [$+2.0^\circ, -24.9^\circ$] & [$+10.67^\circ, -30.67^\circ$] & [$+15.0^\circ, -15.0^\circ$]\\
  \hline
  Angular Resolution (Vertical)            & $0.4^\circ$ & $1.33^\circ$ & $2.0^\circ$\\
  \hline
  Points/Sec in Millions        & $\sim1.3$ & $\sim0.7$  & $\sim0.3$  \\
  \hline
  Range        & 120m & 100m  & 100m  \\
  \hline
  Noise   & $\pm2.0$cm & $\pm2.0$cm  & $\pm3.0$cm \\
  \hline
\end{tabular}
\caption{Characteristics of different \lidars from \protect\cite{velodyne_wiki}. The prices are representative.}
\label{tab:lidar-configurations}
\end{table*}

\begin{figure*}[h!]
    \centering
    \includegraphics[width=0.8\linewidth]{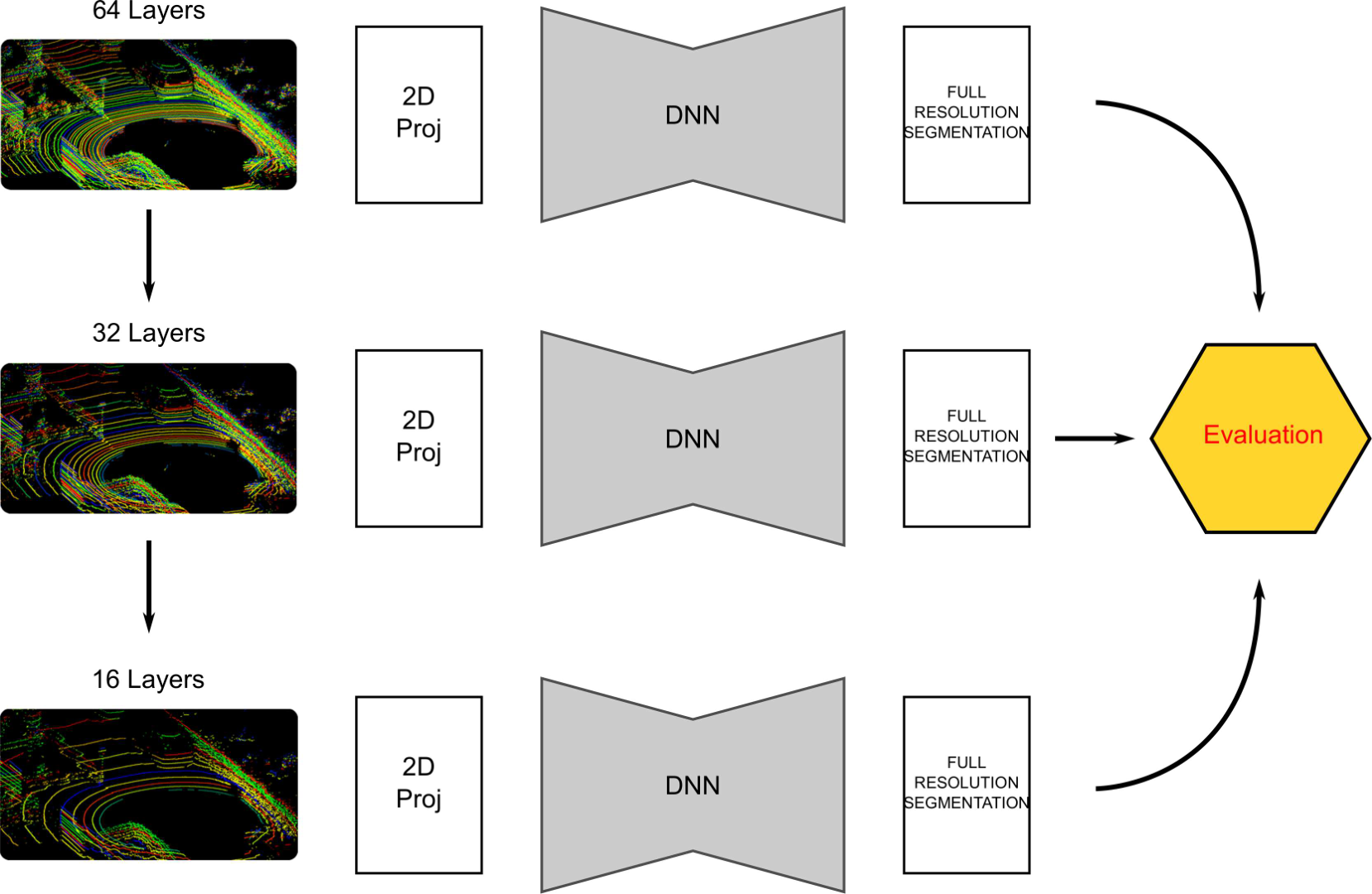}
    \caption{Overall methodology to evaluate the performance of road segmentation across different resolutions. See Figures \ref{fig:all} for more details on the architectures used.}
    \label{fig:pipeline}
\end{figure*}

The objective of this study is to examine the effect of reducing spatial resolution of \lidars by 
subsampling a 64-scanning layers \lidar on the task of road segmentation. This is done to simulate the 
evaluation of low resolution scanners for the task of road segmentation without requiring any pre-existing
datasets on low resolution scanners. The key contribution and goal of our experiment are: First, to 
evaluate the impact of the point cloud's spatial resolution on the quality of the road segmentation task. 
Secondly, determine the effect of subsampling on different point cloud representations, namely on the Bird
Eye View (BEV) and Spherical View (SV), for the task of road segmentation. For BEV representation we use 
existing LoDNN architecture \cite{lodnn2017}, while for SV we use a simple U-net architecture.
In Fig. \ref{fig:pipeline}, we demonstrate a global overview of the methodology used. Finally, we propose
to use surface point normals as complementary feature to the ones already used in current state of the art
research. Results are reported on the KITTI road segmentation benchmark \cite{Fritsch2013ITSC}, and the 
newly introduced Semantic KITTI dataset \cite{behley2019iccv}.

\subsection{Related Work}
\label{subsec:relatedwork}

LoDNN (\lidar Only Deep Neural Networks) \cite{lodnn2017} is a FCN (Fully Convolution Network) based binary segmentation architecture, with encoder containing sub-sampling layers, and decoder with up-sampling layers. The architecture is composed of a core context module that performs multi-scale feature aggregation using dilated convolutions. In the class of non-deep learning methods, authors in \cite{chen2017lidar} built a depth image in spherical coordinates, with each pixel indexed by set of fixed azimuth values ($\phi$) and horizontal polar angles ($\theta$), with intensity equal to the radial distances ($r$). Authors assume for a given scanner layer (a given $\phi$) all points belonging to the ground surface shall have the same distance from the sensor along the $x$ axis.

Authors in \cite{lyu2018chipnet} propose a FCN based encoder-decoder architecture with a branched 
convolutional block called the ChipNet block. It contains filters with ($1 \times 1 \times 64 \times 64$, 
$3 \times 3 \times 64 \times 64$, $3 \times 3 \times 64 \times 64$) convolutional kernels in parallel. 
They evaluate the performance of road segmentation on Ford dataset and KITTI benchmark on a FPGA platform. 
The work closest to our study is by authors  \cite{del2017low}, where they compare a high-resolution 
64-layer \lidar with a low-resolution system, 16-layer \lidar, for the task of vehicle detection. They 
obtain the low resolution 16-layer scans by sub-sampling the $64$-layer scans. The results demonstrate 
that their DNN architecture on low resolution is able outperform their geometric baseline approach. They 
also show similar tracking performance w.r.t their high-resolution HDL-64 sensor at close range.

Additionally \cite{jaritz2018sparse} studies joint sparse-to-dense depth map completion and semantic 
segmentation using NASNet architectures. They work with varying densities of points reprojected into the 
Front View (FV) image, that is the image domain of the camera sensor. Authors achieve an efficient 
interpolation of depth to the complete FOV using features extracted using early and late fusion from the 
RGB-image stream.

\section{Methodology}

A point cloud is a set of points $\{\mathbf{x}_k\}_{k=1}^N \in \mathbb{R}^3$. It is usually represented in the cartesian coordinate system where the 3-dimensions correspond to the $(x, y, z)$. A \lidar scan usually consists of a set of such $3$D-points obtained with the sensor as origin. In this paper each  \lidar scan is projected on an image. The two projections we study are the Bird Eye View (BEV) and Spherical View (SV).

The BEV image is a regular grid on the $x,y$ plane on to which each point is projected. Each cell of the grid corresponds to a pixel of the BEV image. As in \cite{lodnn2017} we define a grid of $20$ meters wide, $y \in [-10, 10]$, and $40$ meters long, $x \in [6, 46]$. This grid is divided into cells of size $0.10 \times 0.10$ meters. Within each cell we evaluate six features: number of points, mean reflectivity, and mean, standard deviation, minimum, and maximum elevation. Each point cloud is thus projected and encoded in a tensor of $400 \times 200 \times 6$,  where $400, 200$ are the BEV image height and width. We refer to the set of these  six features as \emph{BEV Classical Features} (see Fig. \ref{fig:bev-features}). 

\begin{figure}[h!]
    \centering
    \includegraphics[width=\columnwidth]{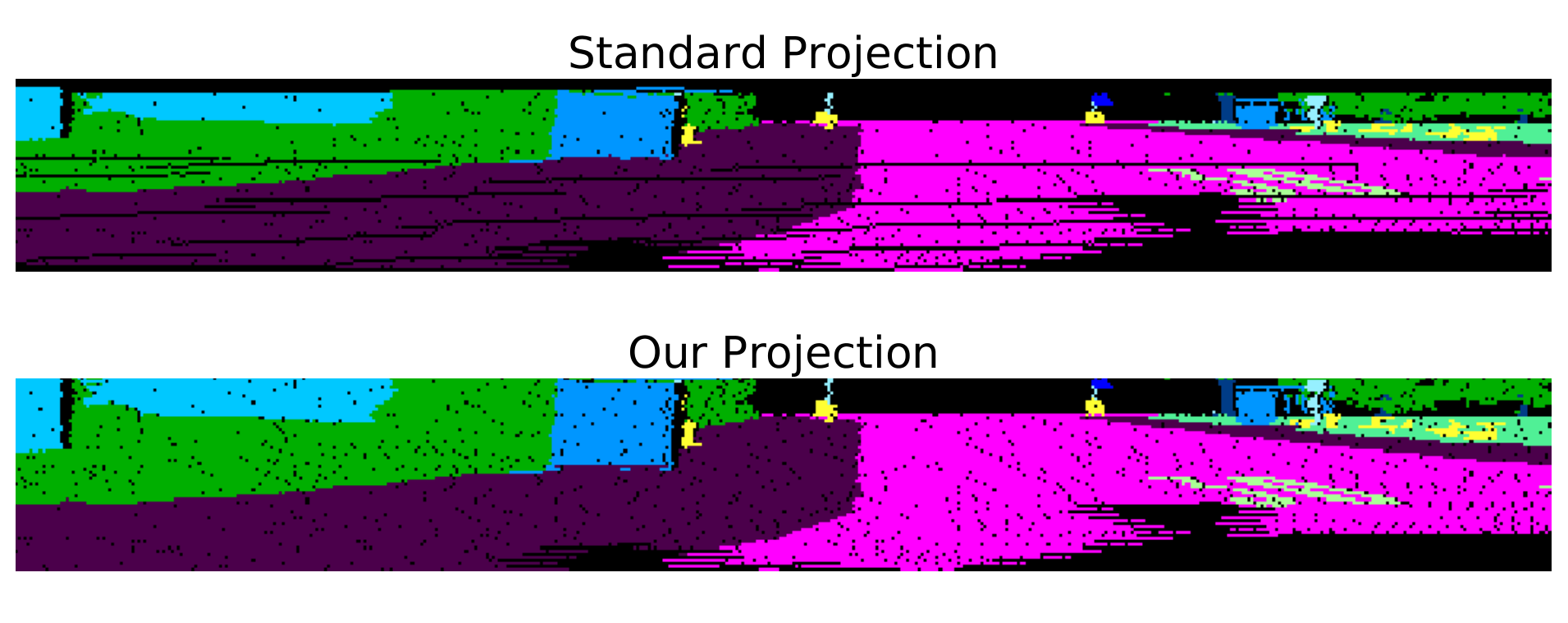}
    \caption{SV projection: two cropped images showing difference between the standard projection, and our projection.}
    \label{fig:sv-vs-our}
\end{figure}

In SV image, each point $\mathbf{x} = (x,y,z)$ is first represented using spherical coordinates $(\rho, \varphi, \theta)$:

\begin{equation*} \label{eq:projection}
\begin{cases}
  \rho = \sqrt{x^2 + y^2 + z^2}, \\ 
  \varphi = \atantwo(y,z), \\
  \theta = \arccos(z/\rho), \\ 
  \end{cases}
\end{equation*}
and then projected on a grid over the sphere $S^2 = \{x^2 + y^2 + z^2 = 1\}$. The size of the cells in the grid is chosen accordingly to the Field Of View (FOV) of scanner. For instance, \cite{behley2019iccv} project point clouds to $64 \times 2048$ pixel images by varying the azimuth angle ($\theta$) and vertical angle ($\varphi$) into two evenly discretised segments. In our case, we use a slightly different version of the SV. Instead of evenly dividing the vertical angle axis and associating a point to a cell according to its vertical and azimuth angles, we retrieve for each point the scanner layer that acquired it and we assign the cell according to the position of the layer in the laser stack and the value of the azimuth angle. Basically, for a scanner with $64$ layers we assign the point $\mathbf{x}$ to a cell in row $i$ if $\mathbf{x}$ has been acquired by the $i$-th layer starting from the top. We decided to use this approach because in the scanner the layers are not uniformly spaced and using standard SV projection causes that different points collide on the same cell and strips of cells are empty in the final image as illustrated in Fig.\ref{fig:sv-vs-our}. In subsection \ref{subsec:subsampling}, we describe how we associate each point at the layer that captured it. However, we underline that our method relies on the way the points are ordered in the point cloud array. 

Finally, following \cite{velas2018cnn}, in each grid cell we compute the minimum elevation, mean reflectivity and minimum radial distance from the scanner. This information is encoded in a three-channel image. In Fig. \ref{fig:sv-features}, an example of the SV projection is shown in the first three images from the top. Since these three features are already used in the state of the art for the ground segmentation task, in SV we refer to \emph{SV Classical Features}, as the set of SV images composed by minimum elevation, mean reflectivity and minimum radial length.

Once extracted these feature maps we use them as input to train DNNs for binary segmentation. We trained LoDNN model for the case of BEV projection, and U-Net model in the case of SV projection.   

\subsection{DNN models} 
\begin{figure*}[h!]
    \centering
    \begin{subfigure}[t]{\textwidth}
        \centering
        \includegraphics[width=0.85\linewidth]{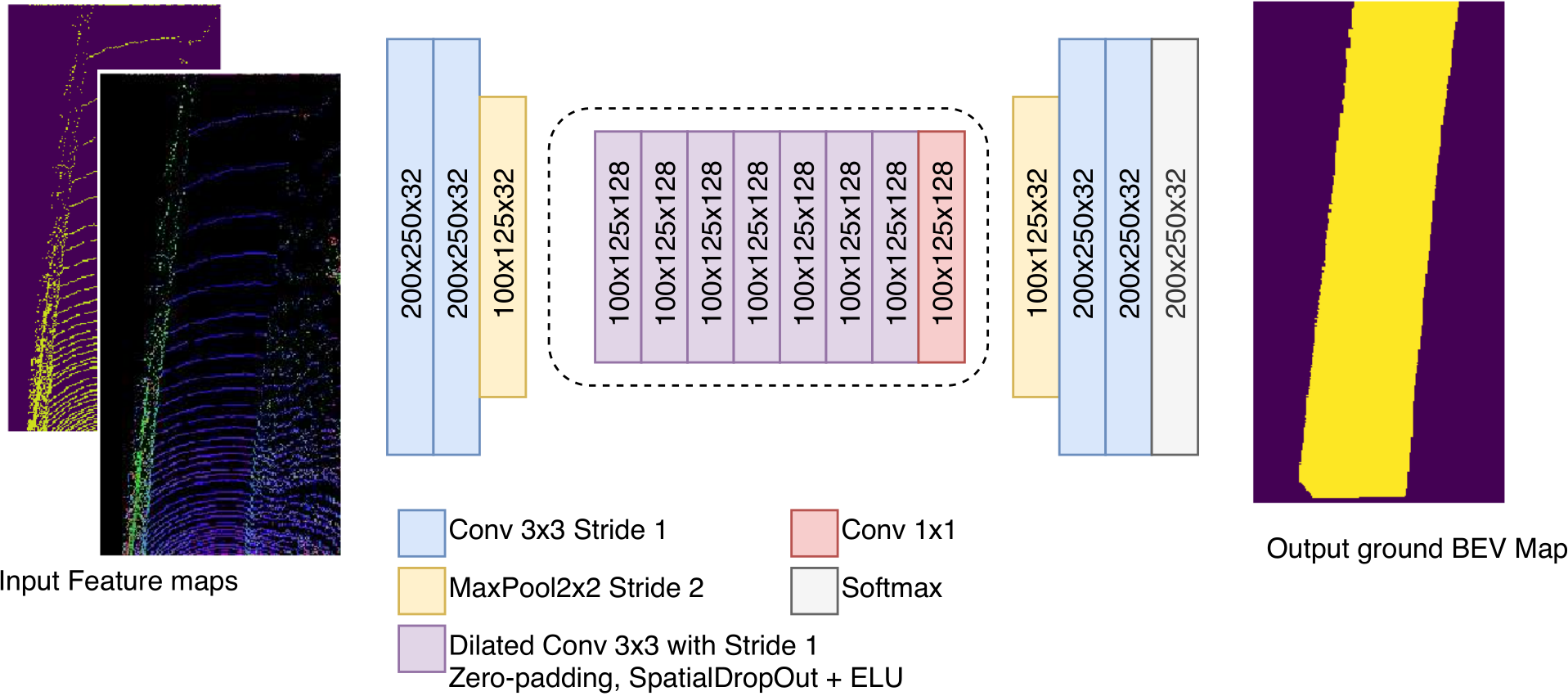}
        \caption{LoDNN Architecture by authors \protect\cite{lodnn2017} in our experiments on BEV.}
        \label{fig:lodnn}
    \end{subfigure}
    \begin{subfigure}[t]{\textwidth}
        \centering
        \includegraphics[height=1.8in]{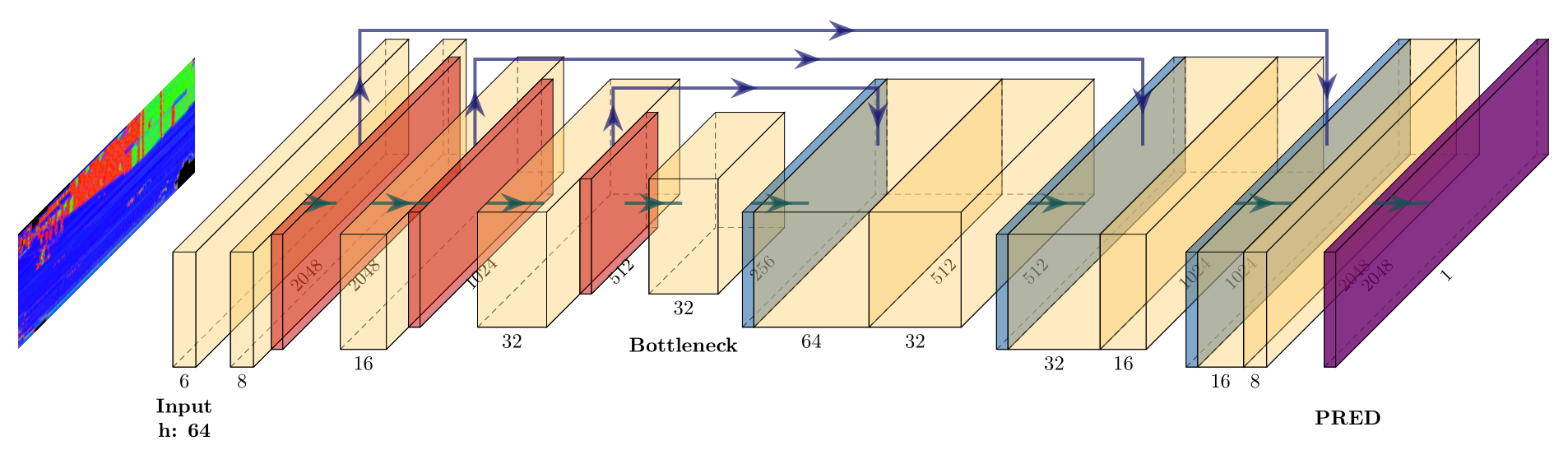}
        \caption{U-Net architecture used for the $64$ layer.}
    \label{fig:unet64}
    \end{subfigure}
    \begin{subfigure}[t]{\textwidth}
        \centering
        \includegraphics[width=\textwidth]{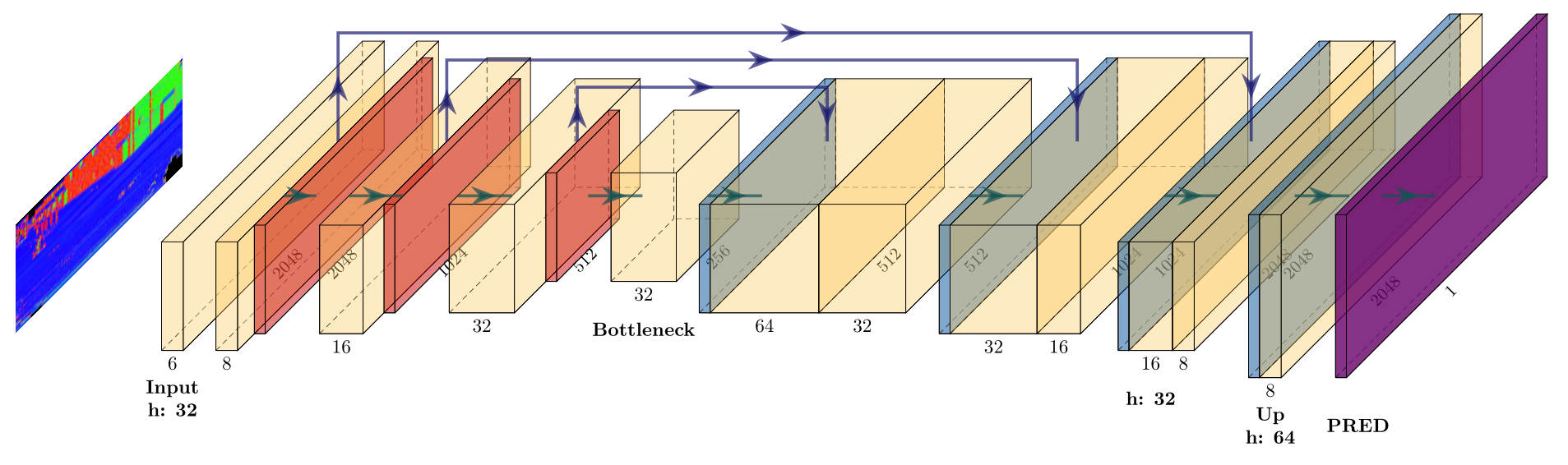}
        \caption{U-Net architecture used for the $32$ layer.}
    \label{fig:unet32}
    \end{subfigure}
    \begin{subfigure}[t]{\textwidth}
        \centering
        \includegraphics[width=\textwidth]{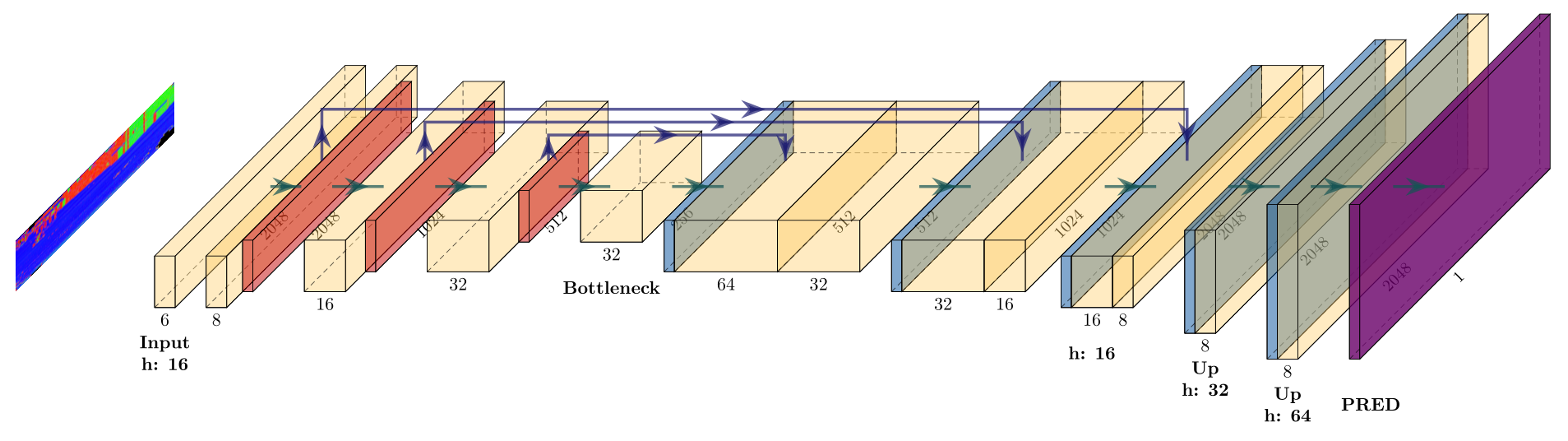}
        \caption{U-Net architecture used for the $16$ layer.}
    \label{fig:unet16}
    \end{subfigure}
    
    \caption{(a) LoDNN Architecture used on BEV images. (b-d) U-Net Architectures used on SV images.}
    \label{fig:all}
\end{figure*}

The LoDNN architecture from \cite{lodnn2017} is a FCN designed for semantic segmentation, it has an input layer that takes as input the BEV images, an encoder, a context module that performs multi-scale feature  aggregation  using  dilated  convolutions and a decoder which returns confidence map for the road.
Instead of Max unpooling layer \footnote{In the context of DNN, a \emph{layer} is a general term that applies to a collection of 'nodes' operating together at a specific depth within a neural network. In the context of \lidar scanners, the \emph{number of scanning layer} refers to the number of laser beams installed in the sensor. } specified in \cite{lodnn2017}, we use a deconvolution layer \cite{zeiler2010deconvolutional}. Other than this modification, we have followed the authors implementation of the LoDNN. The architecture is reported in Fig. \ref{fig:lodnn}.

The U-Net architecture \cite{ronneberger2015unet}, is a FCN designed for semantic segmentation. In our implementation of U-Net, the architecture is made of three steps of downsampling and three steps of upsampling. During the downsampling part $1\times2$ max pooling is used to reduce the features spatial size. In order to compare the different cases ($64/32/16$ scanning layers) among them, the $64$ scanning layers ground truth is used for all the cases. For this purpose an additional upsampling layer of size $2\times 1$ is required at the end of the $32$-based architecture and two upsampling layers at the end of the $16$.  In fact the size of SV images for the $32$ scanning layer is $32\times2048$. Thus without an additional upsampling layer we would obtain an output image whose size is $32\times 2048$.  Similarly, for the $16$ scanning layer, we add two upsampling layers of size $2\times 1$ to go from $16\times2048$ to $64\times2048$ pixels output images. Fig. \ref{fig:unet64}, \ref{fig:unet32} \& \ref{fig:unet16} illustrate the three architectures used.

In both cases a confidence map is generated by each model. Each pixel value specifies the probability of whether corresponding grid cell of the region belongs to the road class. The final segmentation is obtained by thresholding at 0.5 the confidence map.

\subsection{Sub-sampling point clouds to simulate low resolution}\label{subsec:subsampling}

\begin{figure}[th!]
    \centering
    \includegraphics[width=\linewidth]{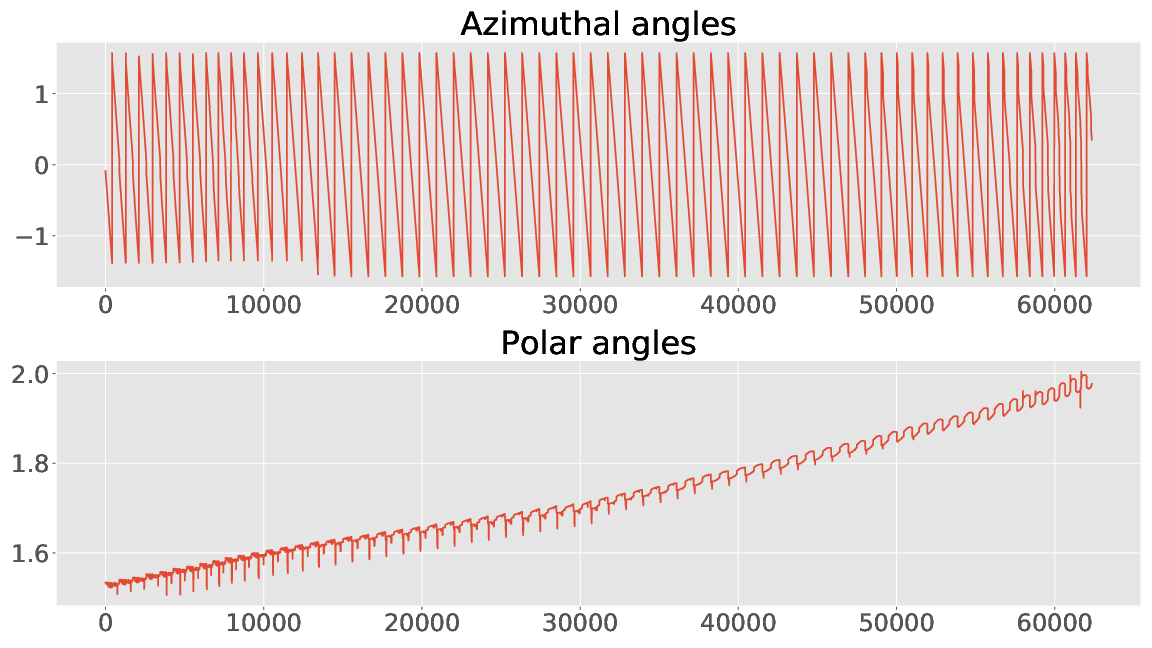}
    \caption{Plot containing the azimuths and vertical angles for a single point cloud.}
    \label{fig:angles}
\end{figure}
Currently there are no dataset available containing scans of the same environment taken simultaneously with scanners at different resolutions, we simulated the 32 and 16 layer scans removing layers from 64 scans. To achieve this, we first need to associate each point within each 64 scan to the scanning layer it was captured from. We exploit the special ordering in which point clouds have been stored within the KITTI datasets. Fig. \ref{fig:angles} shows the azimuth and polar angles of the sequence of points of a single scan. We observe that the azimuth curve contains $64$ cycles over $2\pi$ degrees, while polar curve globally increase. Thus a layer corresponds to a round of $2\pi$ degrees in the vector of azimuths. Scanning layers are stored one after another starting from the uppermost layer to the lowermost one. As we step through sequentially $2\pi$ in the azimuth angle (two zero crossings), we label each point to be within the same layer.
Once retrieved the layers, we can obtain a 32 scan removing one layer out of two from the 64 scan, and obtain a 16 scan removing three layers out of four. The size of SV input images changes when we remove layers. We move from $64\times2048$ pixels for a $64$ layer scanner to $32\times2048$ pixels for the $32$ layer and to $16\times2048$ pixels for the $16$ layer.

\subsection{Surface normal extraction}

\begin{figure}[t!]
    \centering
    \includegraphics[width=\linewidth]{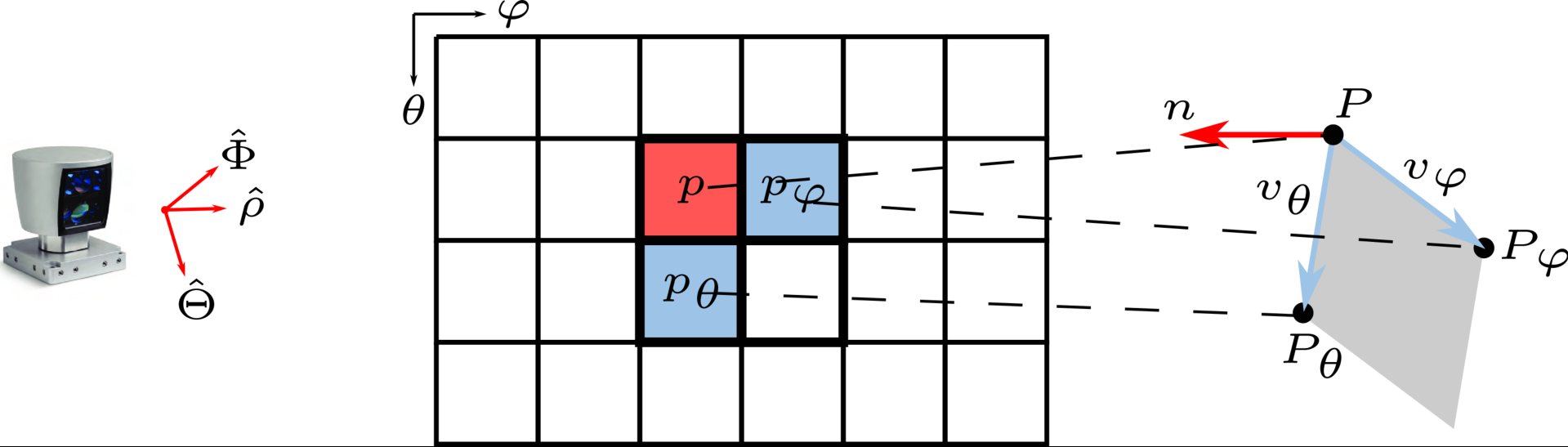}
    \caption{Relationship between adjacent pixels in the radial distance image $\image$ and adjacent points in the $3$D space. Pixels $p$, $p_\varphi$ and $p_\theta$ are associated to $3$D points $P$, $P_\varphi$ and $P_\theta$. Since $P$, $P_\varphi$ and $P_\theta$ compose a local plane, we compute their $3$D gradients as tangent vectors $v_\varphi$, $v_\theta$ from a radial distance value at $p$, $p_\varphi$ and $p_\theta$.}
    \label{fig:est_normals}
\end{figure}
Along with features used in the state of the art, we estimate surface normals from the image containing radial distances in the SV. Our method is inspired by the work of \cite{nakagawa2015estimating} where the authors estimate surface normals using depth image gradients. Let $p = (\varphi,\theta)$ a couple of angles in the spherical grid and let $R$ be the image containing the radial distances. We can associate to $p$ a point $P$ in the $3$D space using the formula
\begin{equation} \label{eq:back-projection}
 \Psi(\varphi, \theta) = \begin{cases}
  x = \image(\varphi,\theta)\cos(\varphi)\sin(\theta), \\ 
  y = \image(\varphi,\theta)\sin(\varphi)\sin(\theta), \\
  z = \image(\varphi,\theta)\cos(\theta). \\ 
  \end{cases}
\end{equation}
Now, let $p_\varphi = (\varphi + \Delta\varphi,\theta)$ and $p_\theta = (\varphi,\theta  + \Delta\theta)$ respectively the vertical and the horizontal neighbouring cells. They have two corresponding points $P_\varphi$ and $P_\theta$ in the $3$D space, as well. Since $P$, $P_\varphi$ and $P_\theta$ compose a local $3$D plane, we can estimate the normal vector $\frac{\partial \Psi}{\partial \varphi} \times \frac{\partial \Psi}{\partial \varphi}$ at $P$ using the two vectors $v_\varphi, v_\theta$ spanning the local surface containing $P$, $P_\varphi$ and $P_\theta$, as in Fig. \ref{fig:est_normals}. We compute $v_\varphi$ using the values of the radial distance image $\image$ at pixels $p$, $p_\varphi$ as
\[
 v_\varphi(\varphi,\theta) = \begin{pmatrix}
  d_\varphi\image(\varphi, \theta)\cos(\varphi)\sin(\theta) + \image(\varphi,\theta)\cos(\varphi)\cos(\theta) \\ 
  d_\varphi\image(\varphi, \theta) \sin(\varphi) \sin(\theta) + \image(\varphi,\theta)\sin(\varphi)\cos(\theta)  \\ 
  d_\varphi \image(\varphi, \theta) \cos(\theta) - \image(\varphi,\theta)\sin(\theta) 
 \end{pmatrix}
\]

where 
\begin{equation}
\begin{aligned}
d_\varphi \image(\varphi,\theta) = & \frac{\image(\varphi+\Delta\varphi,\theta)-\image(\varphi,\theta)}{\Delta\varphi} \\
      = & \frac{\image(p_\varphi)-\image(p)}{\Delta\varphi} \approx \frac{\partial \image}{\partial \varphi}(\varphi, \theta).
\end{aligned}
\end{equation}

Similarly $v_\theta$ is obtained using values at $p$ and $p_\theta$ as:
\[
 v_\theta(\varphi,\theta) = \begin{pmatrix}
d_\theta\image(\varphi, \theta)\cos(\varphi)\sin(\theta) - \image(\varphi,\theta)\sin(\varphi)\sin(\theta) \\ 
  d_\theta\image(\varphi, \theta) \sin(\varphi)\sin(\theta) + \image(\varphi,\theta) \cos(\varphi)\sin(\theta)  \\ 
  d_\theta\image(\varphi, \theta) \cos(\theta) 
 \end{pmatrix}
\]
where 
\begin{equation}
\begin{aligned}
d_\theta\image(\varphi, \theta) = & \frac{\image(\varphi,\theta+\Delta\theta)-\image(\varphi,\theta)}{\Delta\theta} \\
      = & \frac{\image(p_\theta)-\image(p)}{\Delta\theta} \approx \frac{\partial \image}{\partial \theta}(\varphi, \theta).
\end{aligned}
\end{equation}

\begin{figure}[h!]
    \centering
    \includegraphics[width=\columnwidth]{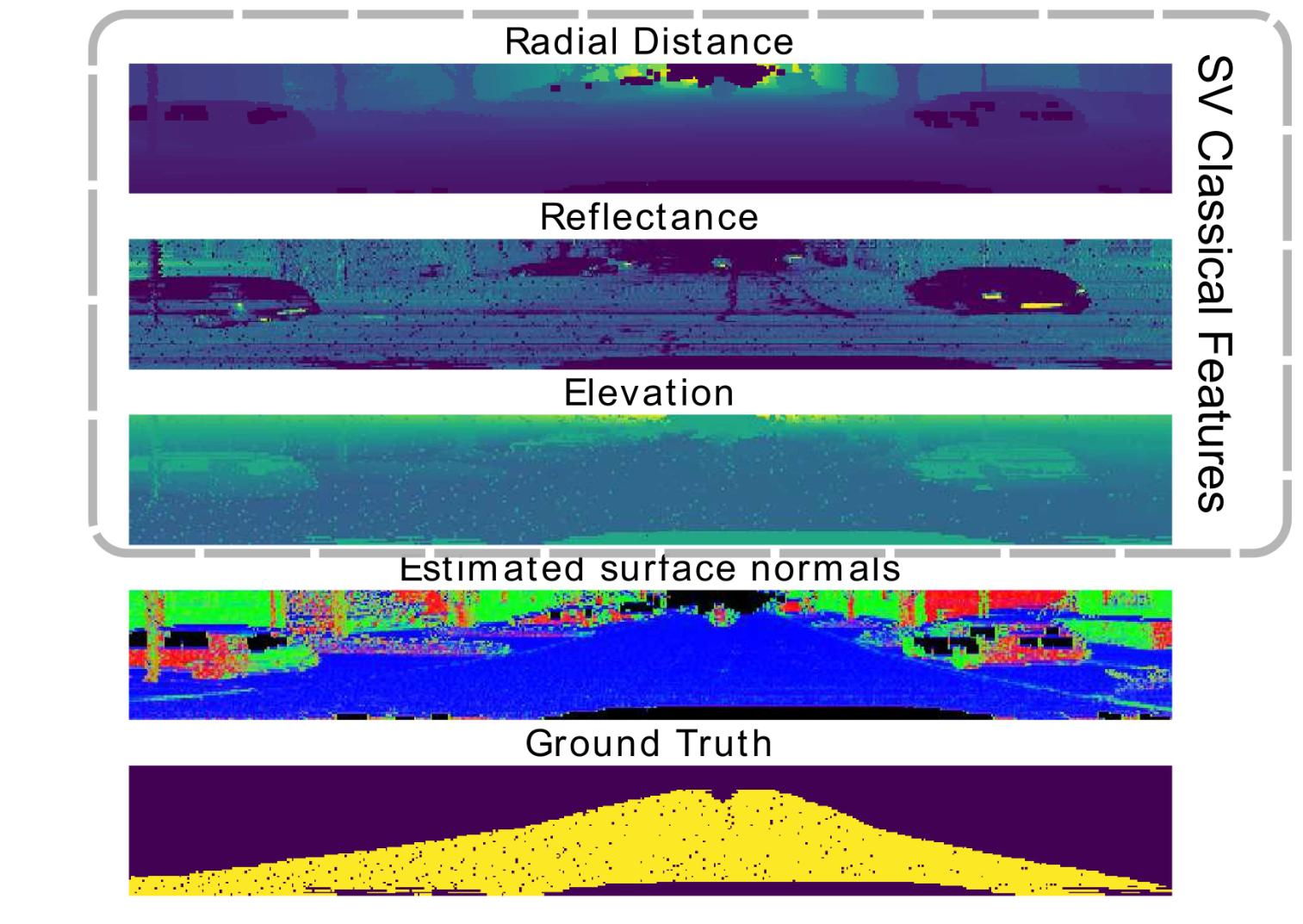}
    \caption{A crop example of features projected on the SV in case of a $64$ layers scanner. Surface normals are estimated from radial distance image. The last image below is the ground truth for this case.}
    \label{fig:sv-features}
\end{figure}

The approximated normal vector is $n = v_\varphi \times v_\theta$. Once the surface point normals are estimated in the SV, we get them back to $3$D-cartesian coordinates, and subsequently project them onto the BEV. This adds three supplementary channels to the input images. Fig. \ref{fig:sv-features} shows the results obtained on SV image with $64$ layer. For each pixel we mapped the coordinates $(x, y, z)$ of the estimated normals to the RGB color map, so $x\rightarrow R$, $y\rightarrow G$ and $z\rightarrow B$. Please remark that a FCN can not extract this kind of features through convolutional layers starting from \emph{SV classic features}. In fact, to extract this kind of information using convolution the FCN should be aware of the position of the pixel inside of the image, \emph{i.e.} know the angles $(\varphi, \theta)$, but this would break the translational symmetry of convolutions. This enforces prior geometrical information to be encoded in the features maps that are the input of the DNN.
Finally, we also remark that the normal feature maps are computationally efficient since its a purely local operation in the spherical coordinates.

\begin{figure}[t!]
    \centering
    \includegraphics[width=\columnwidth]{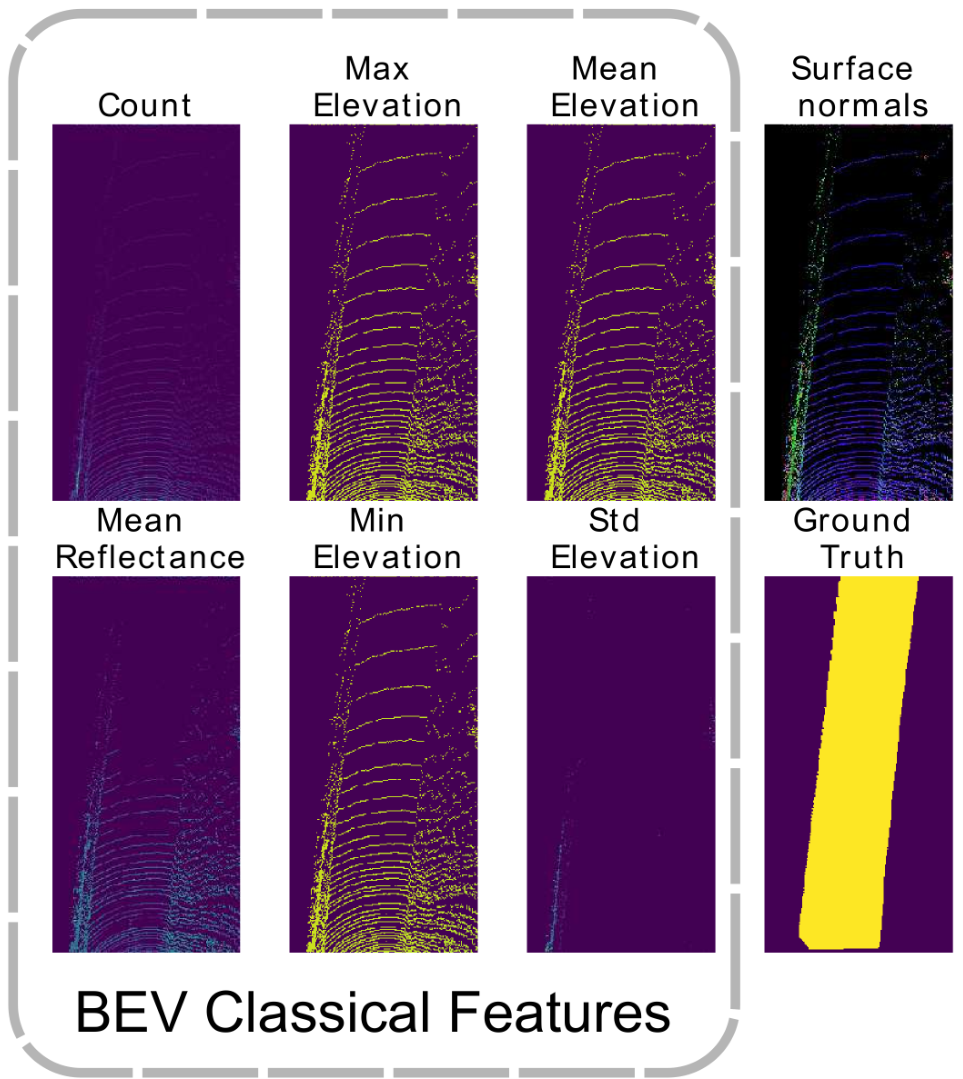}
    \caption{An example of features projected on the BEV in case of a $64$ layers scanner. Surface normals are computed on SV and projected to BEV.}
    \label{fig:bev-features}
\end{figure}

\section{Experiments \& Analysis}

The binary road segmentation task is evaluated on two datassets : 1. the KITTI road segmentation \cite{Fritsch2013ITSC}, 2. Semantic-KITTI dataset \cite{behley2019iccv}. 
The input information to the DNNs comes purely from point clouds and no camera image information was used.
The BEV and SV representations were used over the point clouds from KITTI road-dataset, while only the SV representation over Semantic-KITTI. 
The BEV ground truth information for semantic-KITTI did not currently exist during the redaction of this article, and thus no evaluation was performed. The projection of the $3$D labels to BEV image in semantic-KITTI produced sparsely labeled BEV images and not a dense ground truth as compared the BEV ground truth in Fig. \ref{fig:bev-features}. The SV ground truth images have been generated by projecting $3$D labels to $2$D pixels. We consider a pixel as road if at least one road $3$D point is projected on the pixel.

\textbf{Training} : Adam optimiser with initial learning rate of 0.0001 is used to train the models. The models were trained using an Early-Stopping. We used the Focal loss with gamma factor of $\gamma=2$ \cite{lin2017focal}, thus the resulting loss is $L(p_t) = -(1-p_t)^\gamma \log(p_t)$, where $$p_t=\begin{cases}
               p  & \text{if } y = 1,\\
            1-p & \text{otherwise.}\\
            \end{cases}$$
The focal loss was useful in the KITTI Road segmentation benchmark. The road class was measured to be around 35\% of the train and validation set, in the BEV, while around 5\% for the SV. 
This drastic drop in the road class in SV is due to the restriction of the labels to the camera FOV. While for Semantic-KITTI we observed lesser level of imbalance between road and background classes.

\textbf{Metrics} : We use the following scores to benchmark our experiments. The \Fone-Score and Average Precision are defined as 

\begin{equation} 
\label{eqn:F1-AP}
 F_1= 2 * \frac{P * R}{P + R}, \ \ AP=\sum_n P_n(R_n - R_{n-1})
\end{equation}

where, $P=\frac{TP}{TP + FP} \ \ R=\frac{TP}{TP + FN}$ and $P_n$, $R_n$ are precision and recall at $n$-th threshold. In all the cases the scores have been measured on the projected images, and we report them in Table \ref{tab:resultsKITTI}.

\textbf{Evaluating different resolutions}:
When subsampling point clouds, the input SV image size changes accordingly. For example, after the first subsampling the input SV image now has a size of $32 \times 2048$. In order to get fair comparison, the evaluation of all results is made at the original full resolution at $64 \times 2048$. In such as case the number of layers in the U-Net architectures has been increased to up-sample the output segmentation map to the full resolution. This lead to 3 different architectures for 16, 32 and 64 layers, see Fig. \ref{fig:unet64}, \ref{fig:unet32} \& \ref{fig:unet16}. Three different models were trained on the different SV images. In the Semantic KITTI dataset the evaluation has been done over the road class. The BEV image on the other hand remains the same size with subsampling. Though subsampling in BEV introduces more empty cells as certain layers disappear. 

\subsection{KITTI road estimation benchmark} 
The KITTI road segmentation dataset consists of three categories: urban unmarked (UU),urban marked (UM), and urban multiple marked lanes (UMM).
Since the test dataset's ground truth is not publicly available, 289 training samples from the dataset is split into training, validation and test sets for the experiments. Validation and test sets have 30 samples each and the remaining 229 samples are taken as training set.

Ground truth annotations are represented only within the 
camera perspective for the training set. We use the ground truth annotations provided by 
authors \cite{lodnn2017} in our experiments. The BEV groudtruth was generated over 
the xy-grid within $[-10,10]\times[6,46]$ with squares of size $0.10 \times 0.10$ meters. 

\begin{figure}[h!]
    \centering
    \includegraphics[width=\linewidth]{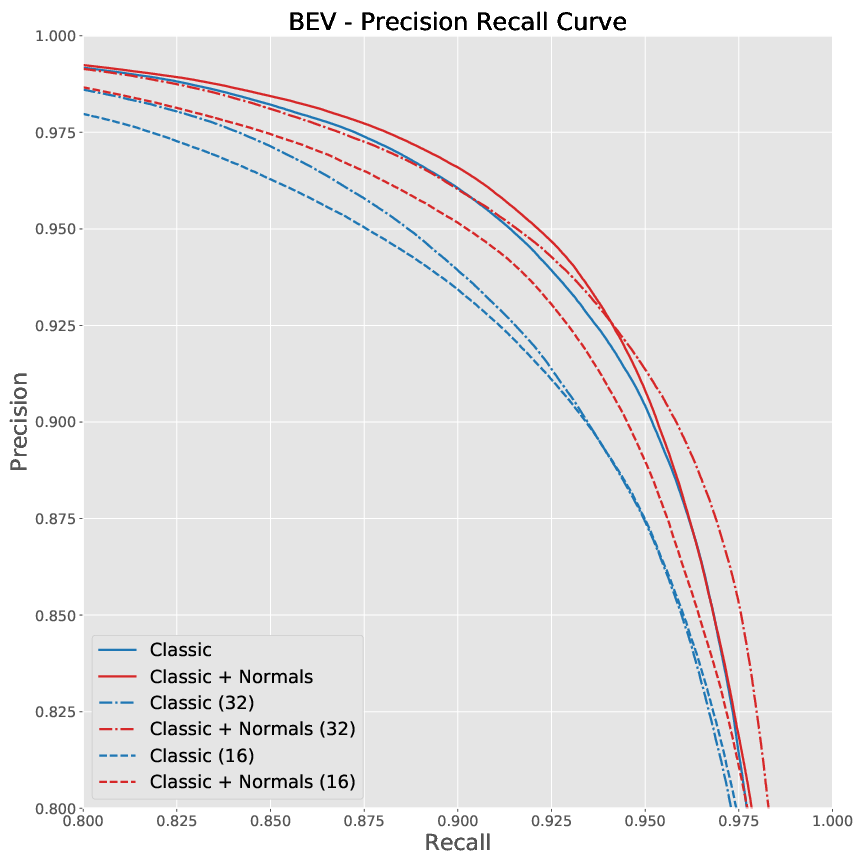}
    \caption{KITTI Road Segmentation with BEV images: Precision-Recall Curve for various features with and without sub-sampling.}
    \label{fig:kitti-bev-roc}
\end{figure}

\begin{figure}[h!]
    \centering
    \includegraphics[width=\linewidth]{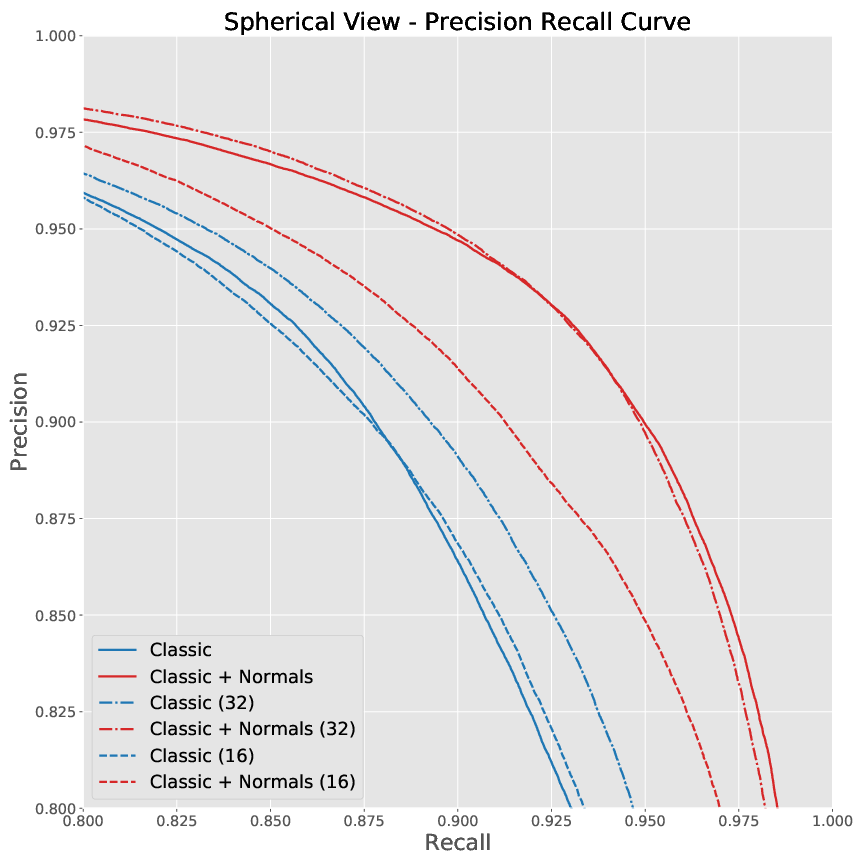}
    \caption{KITTI Road Segmentation with SV images: Precision-Recall Curve for various features with and without sub-sampling.}
    \label{fig:kitti-sv-roc}
\end{figure}

Figures \ref{fig:kitti-bev-roc}, \ref{fig:kitti-sv-roc} illustrate the Precision-Recall 
(PR) curves obtained on BEV images and SV images. 
The performance metrics for the different resolutions (64/32/16 scanning layers) of the scanners are reported, 
for both the classical and classical-with-normal features. 
At full resolution the classic features obtain state of the art scores as reported by authors \cite{lodnn2017}. 
In Table \ref{tab:resultsKITTI}, we observe that with subsampling and reduction in the number of layers, 
there is a degradation in the AP along with metrics. With the addition of the normal features, we observe
and improvement in AP across all resolutions/number of layers. 

\begin{table}[h!]
\centering
\begin{tabular}{|m{3.3cm}|m{0.70cm}|m{0.70cm}|m{0.70cm}|m{0.70cm}|} 
  \hline
  {\textbf{KITTI Road-Seg, BEV}} & \textbf{AP} & \textbf{\Fone} & \textbf{Rec} & \textbf{Prec} \\
  \hline
  Classical (64)            & 0.981 & 0.932 & 0.944 & 0.920\\
  Classical + Normals (64)  & 0.983 & 0.935 & 0.945 & 0.926 \\
  Classical (32)            & 0.979 & 0.920 & 0.926 & 0.914\\
  Classical + Normals (32)  & 0.984 & 0.934 & 0.937 & 0.930\\
  Classical (16)            & 0.978 & 0.918 & 0.920 & 0.915\\
  Classical + Normals (16)  & 0.981 & 0.927 & 0.936 & 0.919\\
\hline
\textbf{KITTI Road-Seg, SV} & \textbf{AP} & \textbf{\Fone} & \textbf{Rec} & \textbf{Prec} \\
  \hline
  Classical (64)            & 0.960 & 0.889 & 0.914 & 0.889\\
  Classical + Normals (64)  & 0.981 & 0.927 & 0.926 & 0.929\\
  Classical (32)            & 0.965 & 0.896 & 0.915 & 0.878\\
  Classical + Normals (32)  & 0.981 & 0.927 & 0.928 & 0.927\\
  Classical (16)            & 0.960 & 0.888 & 0.900 & 0.875\\
  Classical + Normals (16)  & 0.974 & 0.906 & 0.914 & 0.899\\
\hline
\end{tabular}
\caption{Results obtained on the test set of the KITTI road segmentation dataset in the BEV and SV.}
\label{tab:resultsKITTI}
\end{table}

\subsection{Semantic-KITTI}
The Semantic-KITTI dataset is a recent dataset that provides a pointwise label across the different sequences from KITTI Odometry dataset, for various road scene objects, road, vegeation, sidewalk and other classes. The dataset was split into train and test datasets considering only the road class. To reduce the size of the dataset, and temporal correlation between frames, we sampled one in every ten frames over the sequences 01-10 excluding the sequence 08, over which we reported the our performance scores. The split between training and test has been done following directives in \cite{behley2019iccv}.

\begin{figure}[h!]
    \centering
    \includegraphics[width=\linewidth]{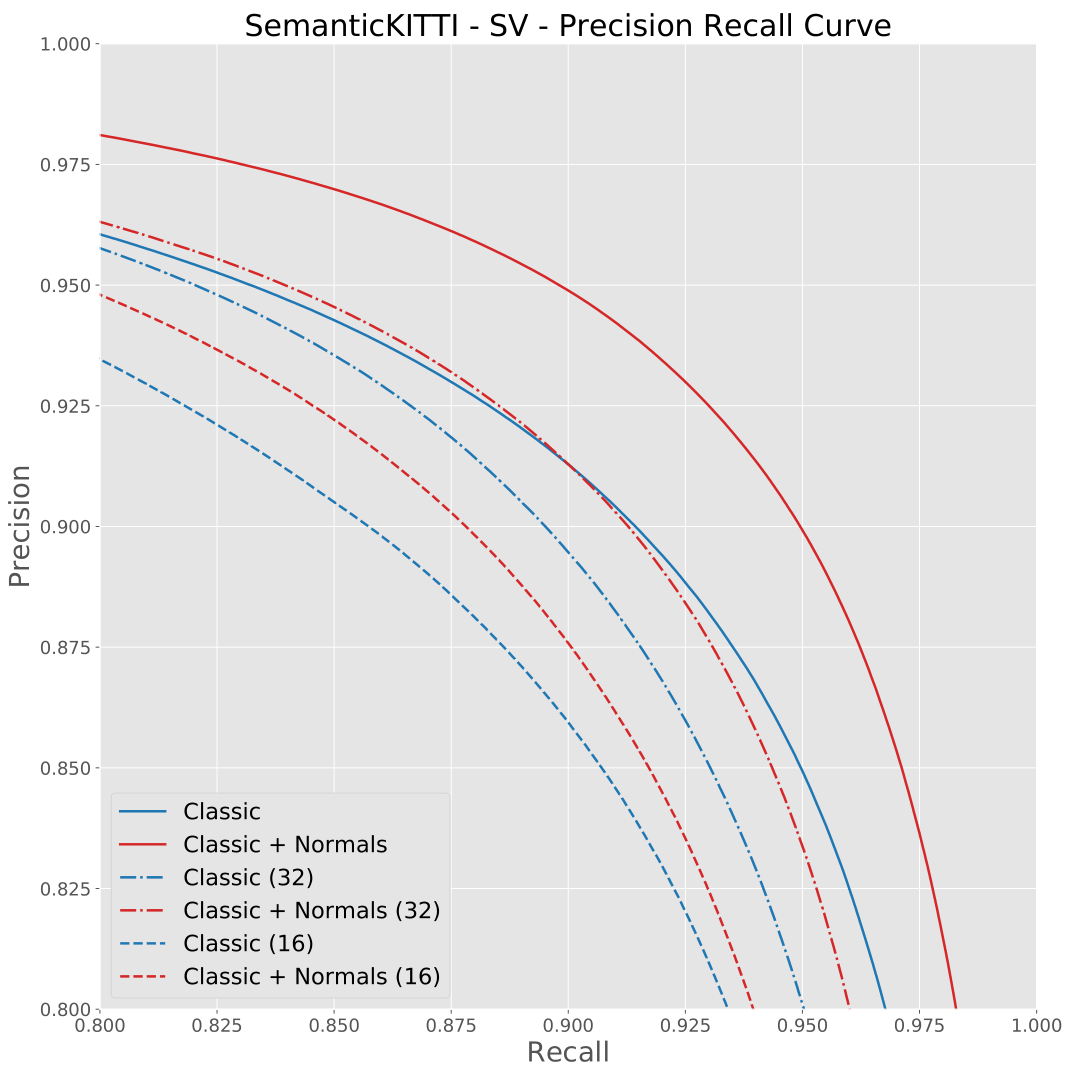}
    \caption{SemanticKITTI with SV images: Precision-Recall Curve for various features with and without sub-sampling.}
    \label{fig:sk-sv-roc}
\end{figure}

With the decrease in vertical angular resolution by subsampling the original 64 layer SV image we observe a minor but definite drop in the binary road segmentation performance (in various metrics)
for sparse point clouds with 32 and 16 scanning layers. This is decrease is visible both in Table \ref{tab:results-SemKITTI} but also in the Precision-Recall curve in Fig. \ref{fig:sk-sv-roc}. With the addition of our normal features to the classical features we do observe a clear improvement in performance across all resolutions (16, 32 and 64 scanning layers). Geometrical normal features channel as demonstrated in Fig. \ref{fig:sv-features} show their high correlation w.r.t the road class region in the ground-truth. Road and ground regions represent demonstrate surfaces which are low elevation flat surfaces with normal's homogeneously pointing in the same directions.

\begin{table}[h!]
\centering
\begin{tabular}{|m{3.3cm}|m{0.70cm}|m{0.70cm}|m{0.70cm}|m{0.70cm}|} 
  \hline
  \textbf{Semantic KITTI-SV} & \textbf{AP} & \textbf{\Fone} & \textbf{Rec} & \textbf{Prec} \\
  \hline
  Classical (64)            & 0.969 & 0.907 & 0.900 & 0.914\\
  Classical + Normals (64)  & 0.981 & 0.927 & 0.927 & 0.927\\
  Classical (32)            & 0.958 & 0.897 & 0.902 & 0.892\\
  Classical + Normals (32)  & 0.962 & 0.906 & 0.906 & 0.906\\
  Classical (16)            & 0.944 & 0.880 & 0.879 & 0.882\\
  Classical + Normals (16)  & 0.948 & 0.889 & 0.894 & 0.883\\
  \hline
\end{tabular}
\caption{Results obtained on the test set of the Semantic-KITTI dataset in the SV.}
\label{tab:results-SemKITTI}
\end{table}

\section{Conclusion}
In view of evaluating the performance of low-resolution \lidars for segmentation of the road class, in this study we evaluate the effect of subsampled \lidar point clouds on the performance of prediction. This is to simulate the evaluation of low resolution scanners for the task of road segmentation. As expected, reducing the point cloud resolution reduces the segmentation performances. Given the intrinsic horizontal nature of the road we propose to use estimated surface normals. These features cannot be obtained by a FCN using \emph{Classical Features} as input. We demonstrate that the use of normal features increase the performance across all resolutions and mitigate the deterioration in performance of road detection due to subsampling in both BEV and SV. Normals features encode planarity of roads and is robust to subsampling.

\subsection{Future work}
In future works we aim to study the effect of temporal aggregation of \lidar scans on reduced spatial resolution due to subsampling the vertical angle. Furthermore, in SV we upsampled the information inside the network, just before the prediction layer. In future works we would like to upsample the input SV range images, and evaluate the performance of the U-Net model on the original $64 \times 2048$ sized SV range image. The up-sampling can be trained end-to-end to achieve successful reconstruction of $64 \times 2048$ image from sub-sampled $32 \times 2048$ or $16 \times 2048$ images.

We have focused our study on road segmentation mainly to limit and understand the effect of subsampling on a geometrically simple case. In our future study we aim to evaluate performance of key classes i.e. cars, pedestrians, to determine the loss in accuracy on account of subsampling of pointclouds. 

\section*{Acknowledgements}
\label{Acknowledgements}
We thank authors of LoDNN for their insights and support.

{
	\begin{spacing}{1.17}
		\normalsize
		\bibliography{root} 
	\end{spacing}
}

\end{document}